\documentclass[10pt]{article}
\usepackage{colacl}
\usepackage{times}
\usepackage{epsfig}
\usepackage{for-edmonds-paper}
\usepackage{avm}\avmoptions{active}

\title{\vspace*{-9ex}{\small In {\it Proceedings Proceedings of the
      AMTA/SIG-IL Second Workshop on Interlinguas, Langhorne,} PA,
    October 1998, pp. 23-30.} \\[7ex]
Translating near-synonyms: Possibilities and preferences in the
  interlingua}

\author{Philip Edmonds \\
  Department of Computer Science \\
  University of Toronto \\
  Toronto, Canada M5S 3G4 \\
  \texttt{pedmonds@cs.toronto.edu}}

\begin{document}
\maketitle


\section{Introduction}

This paper argues that an interlingual representation must explicitly
represent some parts of the meaning of a situation as
\emph{possibilities} (or preferences), not as necessary or definite
components of meaning (or constraints).  Possibilities enable the
analysis and generation of nuance, something required for faithful
translation.  Furthermore, the representation of the meaning of words
is crucial, because it specifies which nuances words can convey in
which contexts.

In translation it is rare to find the exact word that faithfully and
directly translates a word of another language.  Often, the target
language will provide many near-synonyms for a source language word
that differ (from the target word and among themselves) in nuances of
meaning.  For example, the French \w{fournir} could be translated as
\w{provide, supply, furnish, offer, volunteer, afford, bring,} and so
on, which differ in fine-grained aspects of denotation, emphasis, and
style.  (Figures~\ref{provide-note} and~\ref{offer-note} show some
of the distinctions.)  But none of these options may carry the right
nuances to match those conveyed by \w{fournir} in the source text;
unwanted extra nuances may be conveyed, or a desired nuance may be
left out.  Since an exact match is probably impossible in many
situations, faithful translation will require uncovering the nuances
conveyed by a source word and then determining how the nuances can be
conveyed in the target language by appropriate word choices in any
particular context.  The inevitable mismatches that occur are one type
of \defn{translation mismatch}---differences of meaning, but not of
form, in the source and target language \cite{kameyama91}.\footnote{A
  separate class of difference, \defn{translation divergence},
  involves differences in the form of the source and target texts and
  results from lexical gaps in the target language (in which no single
  word lexicalizes the meaning of a source word), and from syntactic
  and collocational constraints imposed by the source language.
  `Paraphrasing' the source text in the target language is required in
  order to preserve the meaning as much as possible
  \cite{dorr94,stede96,elhadad97}.  But even when paraphrasing,
  choices between near-synonyms will have to be made, so, clearly,
  translation mismatches and translation divergences are not
  independent phenomena.  Just as standard semantic content can be
  incorporated or spread around in different ways, so can nuances of
  meaning.}

\begin{figure}
\begin{description}\small\itemsep 0pt
\item[Provide] may suggest foresight and stress the idea of making
  adequate preparation for something by stocking or shipping \ldots
\item[Supply] may stress the idea of replacing, of making up what is
  needed, or of satisfying a deficiency.
\item[Furnish] may emphasize the idea of fitting something or someone
  with whatever is necessary, or sometimes, normal or desirable.
\end{description}\vspace*{-2ex}
\caption{An abridged entry from \textit{Webster's New Dictionary of
    Synonyms}~\cite{gove73}.}
\label{provide-note}
\end{figure}

\begin{figure}
\begin{description}\small\itemsep 0pt
\item[Offer] and \textbf{volunteer} may both refer to a generous
extending of aid, services, or a desired item. Those who
\textit{volunteer} agree by free choice rather than by submission to
selection or command.
\end{description}\vspace*{-2ex}
\caption{An abridged entry from \textit{Choose the Right
    Word}~\cite{hayakawa94}.}
\label{offer-note}
\end{figure}

\section{Near-synonyms across languages}\label{near-syns}

This section examines how near-synonyms can differ within and across
languages.  I will discuss some of the specific problems of lexical
representation in an interlingual MT system using examples drawn from
the French and English versions of the multi-lingual text provided for
this workshop.  

To be as objective as possible, I'll rely on several dictionaries of
synonym discrimination including, for English, \newcite{gove73} and
\newcite{hayakawa94}, and for French, \newcite{bailly70},
\newcite{benac56}, and \newcite{batchelor93}.  Unless otherwise stated,
the information on differences below comes from one of these reference
books.

\def\pair#1#2{\w{#1}\,::\,\w{#2}}

Notation: Below, `\pair{english}{french}' indicates that the pair of
words or expressions \w{english} and \w{french} correspond to one
another in the multi-lingual text (\ie they are apparent translations
of each other).

\subsubsection*{Fine-grained denotational mismatches}

If a word has near-synonyms, then they most likely differ in
fine-grained aspects of denotation.  Consider the following pairs:
\begin{sentence}
\sent\pair{provides}{fournit} 
\sent\pair{provided}{apportaient} 
\sent\pair{provide}{offrir} 
\sent\pair{brought}{fournissait} 
\sent\label{charge}\pair{brought}{se chargeait}
\end{sentence}
These all share the basic meaning of giving or making available what is
needed by another, but each adds its own nuances.  And these are
not the only words that the translator could have used: in English,
\w{furnish, supply, offer,} and \w{volunteer} would have been
possibilities; in French, \w{approvisionner, munir, pourvoir, nantir,
  pr\'esenter,} among others, could have been chosen.  The differences
are complex and often language-specific.  Figures~\ref{provide-note}
and~\ref{offer-note} discuss some of the differences between the
English words, and figures~\ref{fourni-note} and~\ref{offrir-note}
those between the French words.  And this is the problem for translation:
none of the words match up exactly, and the nuances they carry when
they are actually used are context-dependent.  (Also notice that the
usage notes are vague in many cases, using words like `may' and
`id\'ee'.)

Consider this second example:
\begin{msentence}{began}
\msent\pair{began}{amorc\'e}
\msent\pair{began}{commen\c{c}a}
\msent\pair{started}{au d\'ebut}
\end{msentence}
\w{Amorcer} implies a beginning that prepares for something else;
there is no English word that carries the same nuance, but \w{begin}
appears to be the closest match.  \w{Commencer} also translates as
\w{begin}, although \w{commencer} is a general word in French,
implying only that the thing begun has a duration.  In English,
\w{begin} differs from \w{start} in that the latter can imply a
setting out from a certain point after inaction (in opposition to
\w{stop}).

More pairings that exhibit similar fine-grained denotational
differences include these:
\begin{msentence}{broaden}
\msent\pair{broaden}{\'elargir}
\msent\pair{expand}{\'etendre}
\msent\pair{increase}{accro\^{\i}tre}
\end{msentence}
\begin{msentence}{transformation}
\msent\pair{transformation}{passer}
\msent\pair{transition}{transition}
\end{msentence}
\begin{sentence}
\sent\pair{enable}{permettre}
\end{sentence}
\begin{sentence}
\sent\pair{opportunities}{perspectives}
\end{sentence}
\begin{sentence}
\sent\pair{assistance}{assistance}
\end{sentence}

\begin{figure}
\begin{description}\small\itemsep 0pt
\item[Fourni] a rapport \`a la quantit\'e et ce dit de ce qui \`a
  suffisamment ou en abondance le n\'ecessaire.
\item[Muni] et \textbf{arm\'e} sont relatifs \`a l'\'etat d'une chose rendue
  forte ou capable, \w{muni}, plus g\'en\'erale, annon\c{c}ant un
  secours pour faire quoi que ce soit.
\item[Pourvu] comporte un id\'ee de pr\'ecaution et ce dit bien en
  parlant des avantages naturels donn\'es par une sorte de finalit\'e
  \ldots
\item[Nanti,] muni d'un gage donn\'e par un d\'ebiteur \`a son
  cr\'eancier, par ext.\ muni par pr\'ecaution et, absolumment, assez
  enrichi pour ne pas craindre l'avenir.
\end{description}\vspace*{-2ex}
\caption{An abridged entry from \newcite{benac56}.}
\label{fourni-note}
\end{figure}

\begin{figure}
\begin{description}\small\itemsep 0pt
\item[Offrir,] c'est faire hommage d'une chose \`a quelqu'un, en
  manifestant le d\'esir qu'il l'accepte, afin que l'offre devienne un 
  don.
\item[Pr\'esenter,] c'est offrir une chose que l'on tient \`a la main
  ou qui est l\`a sous les yeux et dont la personne peut \`a
  l'instant prendre possession.
\end{description}\vspace*{-2ex}
\caption{An abridged entry from \newcite{bailly70}.}
\label{offrir-note}
\end{figure}

There are two main problems in representing the meanings of these
words.  First, although some of the nuances could be represented by
simple features, such as \g{foresight} or \g{generous}, most of them
cannot because they are complex and have an `internal' structure.
They are concepts that relate aspects of the situation.  For example,
for \w{furnish}, \g{fitting someone with what is necessary} is not a
simple feature; it involves a concept of \g{fitting}, a patient (the
same patient that the overall situation has), a thing that is
provided, and the idea of the necessity of that thing to someone.
Thus, many nuances must be represented as fully-fledged concepts (or
instances thereof) in an interlingua.

Second, many of the nuances are merely suggested or implied, if they
are conveyed at all.  That is, they are conveyed indirectly---the
reader has the license to decide that such a nuance was
unintended---and as such are not necessary conditions for the
definition of the words.  This has ramifications for both the analysis
of the source text and the generation of the target text because one
has to determine how strongly a certain nuance is intended, if at all
(in the source), and then how it should be conveyed, if it can be, in
the target language.  One should seek to translate indirect
expressions as such, and avoid making them direct.  One must also
avoid choosing a target word that might convey an unwanted
implication.  In any case, aspects of word meaning that are indirect
must be represented as such in the lexicon.

\subsubsection*{Coarse-grained denotational mismatches}

Sometimes the translator chooses a target word that is semantically
quite different from the source word, yet still conveys the same basic
idea.  Considering pair~\ref{charge}, above: \w{bring} seems to mean
to carry as a contribution, and \w{se charger} to take responsibility
for.  Perhaps there are no good equivalents in the opposite languages
for these terms, or alternatively, the words might have been chosen
because of syntactic or collocational preferences---they co-occur with
\pair{leadership}{l'administration}, which are not close translations
either.

In fact, the desire to use natural-sounding syntactic and
collocational structures is probably responsible for many of these
divergences.  In another case, the pair \pair{factors}{raisons} occurs
perhaps because the translator did not want to literally
translate the expressions \pair{Many factors contributed to}{Parmi les
  raisons de}.  Such mismatches are outside the scope of this paper,
because they fall more into the area of translation divergences.  (See
\newcite{smadja96} for research on translating collocations.)

\subsubsection*{Stylistic mismatches}

Words can also differ on many stylistic dimensions, but formality is
the most recognized dimension.\footnote{\newcite{hovy88a} suggests
  others including force and floridity, and \newcite{dimarco93}
  suggest concreteness or vividness.  Actually, it seems that the
  French text is more vivid---if a text on banking can be considered
  vivid at all---than the English, using words such as \w{baptis\'ee,
    \'eclatant, contagieux,} and \w{d\'emunis}.}  Consider the
following pairs:
\begin{msentence}{plans}
\msent\pair{plans}{entend bien}
\msent\pair{plan}{envisagent de}
\end{msentence}
While the French words differ in formality (\w{entend bien} is
formal, and \w{envisagent de} is neutral), the same word was chosen in
English.  Note that the other French words that could have been chosen
also differ in formality: \w{se proposent de} has intermediate
formality, and \w{comptent, avont l'intention,} and \w{proj\`etent
  de} are all neutral.

Similarly, in~\ref{began}, above, \w{amorcer} is more formal than
\w{commencer}.  Considering the other near-synonyms: the English
\w{commence} and \w{initiate} are quite formal, as is the French
\w{initier}.  \w{D\'ebuter} and \w{d\'emarrer} are informal, yet both
are usually translated by \w{begin}, a neutral word in English.
(Notice also that the French cognate of the formal English
\w{commence}, \w{commencer}, is neutral.)

Style, which can be conveyed by both the words and the structure of a
text, is best represented as a global property in an interlingual
representation.  That way, it can influence all decisions that are
made.  (It is probably not always necessary to preserve the style of
particular words across languages.)

A separate issue of style in this text is its use of technical or
domain-specific vocabulary.  Consider the following terms used to
refer to the subject of the text:
\begin{msentence}{bank}
\msent\pair{institution}{institution}
\msent\pair{institution}{\'etablissement}
\msent\pair{institution}{association}
\msent\pair{joint venture}{association}
\msent\pair{programme}{association}
\msent\pair{bank}{\'etablissement}
\msent\pair{bank}{banque}
\end{msentence}
In French, it appears that \w{association} must be used to refer to
non-profit companies and \w{\'etablissement} or \w{banque} for their
regulated (for-profit) counterparts.  In English \w{institution},
among other terms, is used for both.  Consider also the following
pairs:
\begin{msentence}{capital}
\msent\pair{seed capital}{capital initial}
\msent\pair{working capital}{fonds de roulement}
\msent\pair{equity capital}{capital social}
\end{msentence}

\subsubsection*{Attitudinal mismatches}

Words also differ in the attitude that they express.  For example, of
\pair{poor}{d\'emunis}, \w{poor} can express a derogatory attitude,
but \w{d\'emunis} (which can be translated as \w{impoverished})
probably expresses a neutral attitude.  Consider also \pair{people of
  indigenous background}{Indiens}.  Attitudes must be included in the
interlingual representation of an expression, and they must refer
to the specific participant(s) about whom the speaker is expressing an
attitude.

\section{Representing near-synonyms}

Before I discuss the requirements of the interlingual representation,
I must first discuss how the knowledge of near-synonyms ought to be
modelled if we are to account for the complexities of word meaning in
an interlingua.  In the view taken here, the lexicon is given the
central role as bridge between natural language and interlingua.

The conventional model of lexical knowledge, used in many
computational systems, is not suitable for representing the
fine-grained distinctions between near-synonyms \cite{hirst95}.  In
the conventional model, knowledge of the world is represented by
ostensibly language-neutral concepts that are often organized as an
ontology.  The denotation of a lexical item is represented as a
concept, or a configuration of concepts, and amounts to a direct
word-to-concept link.  So except for polysemy and (absolute) synonymy,
there is no logical difference between a lexical item and a concept.
Therefore, words that are nearly synonymous have to be linked each to
their own slightly different concepts.  The problem comes in trying to
represent these slightly different concepts and the relationships
between them.  \newcite{hirst95} shows that one ends up with an
awkward proliferation of language-dependent concepts, contrary to the
interlingual function of the ontology.  And this assumes we can even
build a representative taxonomy from a set of near-synonyms to begin
with.

Moreover, the denotation of a word is taken to embody the necessary
and sufficient conditions for defining the word.  While this has been
convenient for text analysis and lexical choice, since a denotation
can be used as an applicability condition of the word, the model is
inadequate for representing the nuances of meaning that are conveyed
indirectly, which, clearly, are not necessary conditions.

\begin{figure*}
\begin{center}
\psfig{file=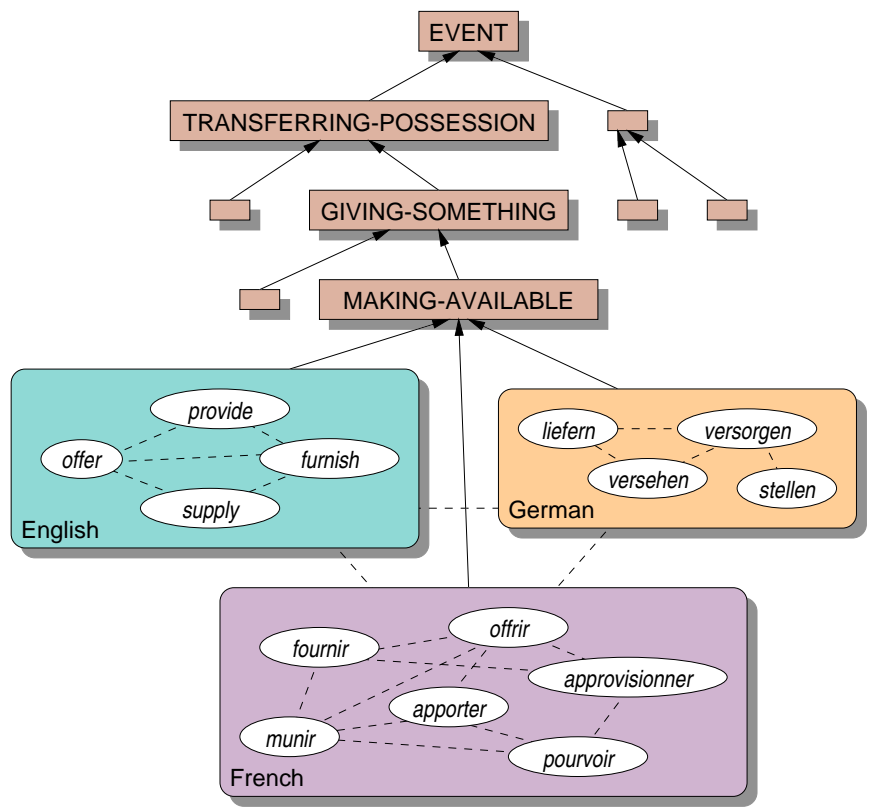,scale=0.833}
\end{center}
\caption{The clustered model of lexical knowledge.}\label{cluster-model}
\end{figure*}

An alternative representation is suggested by the principle behind
Gove's~\shortcite{gove73} synonym usage notes.  Words are grouped into
a entry if they have the same essential meaning, i.e., that they ``can
be defined in the same terms up to a certain point''~(p.\ 25a) and
differ only in terms of minor ideas involved in their meanings.  We
combine this principle with Saussure's paradigmatic view that ``each
of a set of synonyms \ldots\ has its particular value only because
they stand in contrast with one another''~\cite[p.\ 114]{saussure83}
and envision a representation in which the meaning of a word arises
out of a combination of its essential denotation (shared with other
words) and a set of explicit differences to its near-synonyms.

Thus, I propose a \defn{clustered model of lexical knowledge},
depicted in figure~\ref{cluster-model}.  A cluster has two levels of
representation: a core concept and peripheral concepts.  The
\defn{core concept} is a denotation as in the conventional model---a
configuration of concepts (that are defined in the ontology) that
functions as a necessary applicability condition (for choice)---but it
is shared by the near-synonyms in the cluster.  In the figure, the
ontological concepts are shown as rectangles; in this case all three
clusters denote the concept of \textsc{making-available}.  All of the
\defn{peripheral concepts} that the words may differ in denoting,
suggesting, or emphasizing are also represented as configurations of
concepts, but they are explicitly distinguished from the core concept
as indirect meanings that can be conveyed or not depending on the
context.  In the figure, the differences between words (in a single
language) are shown as dashed lines; not all words need be
differentiated.  Stylistic, attitudinal, and collocational factors are
also encoded in the cluster.

Each language has its own set of clusters.  Corresponding clusters
(across languages) need not have the same peripheral concepts since
languages may differentiate their synonyms in entirely different
terms.  Differences across languages are represented, for convenience,
by dashed lines between clusters, though these would not be used in
pure interlingual MT.  Essentially, a cluster is a language-specific
\defn{formal usage note}, an idea originated by \newcite{dimarco93}
that \newcite{edmonds-thesis} is formalizing.

\section{Interlingual representation}

Crucially, an interlingual representation should not be tied to any
particular linguistic structure, whether lexical or syntactic.

Assuming that one has constructed an ontology or domain model (of
language-neutral concepts), an interlingual representation of a
situation is, for us, an instantiation of part of the domain
knowledge.  Both \newcite{stede96} and \newcite{elhadad97} have
developed such formalisms for representing the input to natural
language generation applications (the former to multilingual
generation), but they are applicable to interlingual MT as well.  The
formalisms allow their applications to paraphrase the same input in
many ways including realizing information at different syntactic
ranks and covering/incorporating the input in different ways.  For
them, generation is a matter of satisfying two types of constraints:
(1) covering the whole input structure with a set of word denotations
(thereby choosing the words), and (2) building a well-formed syntactic
structure out of the words.  But while their systems can provide many
options to choose from, they lack the complementary ability to
actually choose which is the most appropriate.

Now, finding the most appropriate translation of a word involves a
tradeoff between many possibly conflicting desires to express certain
nuances in certain ways, to establish the right style, to observe
collocational preferences, and to satisfy syntactic constraints.  This
suggests that lexical choice is not a matter of satisfying constraints
(\ie of using the necessary applicability conditions of a word), but
rather of attempting to meet a large set of \defn{preferences}.  Thus,
a distinction must be made between knowledge that should be treated as
preferences as opposed to constraints in the interlingual
representation.  In the generation stage of MT, one attempts to choose the
near-synonym from a cluster (activated because of the constraints)
whose peripheral concepts best meet the most preferences.

Turning to the analysis stage of MT, since many nuances are expressed
indirectly and are influenced by the context, one cannot know for sure
whether they have been expressed unless one performs a very thorough
analysis.  Indeed, it might not be possible for even a thorough
analysis to decide whether a nuance was expressed, or how indirectly
it was expressed, given the context-dependent nature of word meaning.
Thus, on the basis of the knowledge of what words can express, stored
in the clusters, the analysis stage would output an interlingual
representation that includes \defn{possibilities} of what was
expressed.  The possibilities then become preferences during
generation.

\section{Examples}

Figures~\ref{ex-1}--\ref{ex-last} give examples of interlingual
representations for four segments of the text that involve some of the
words discussed in section~\ref{near-syns}.  Since my focus is on word
meanings, I will not give complete representations of the expressions.
Also note that while I use specific ontological concepts in these
descriptions, this in no way implies that I claim these are the right
concepts to represent---in fact, some are quite crude.  A good
ontology is crucial to MT, and I assume that such an ontology will in
due course be constructed.

I have used attribute-value structures, but any equivalent formalism
would do.  Square brackets enclose recursive structures of
instantiations of ontological concepts.  Names of instances are in
lowercase; concepts are capitalized; relations between instances are
in uppercase; and cross-reference is indicated by a digit in a square.
A whole interlingual representation is surrounded by brace brackets
and consists of exactly one specification of the situation and any
number of possibilities, attitudes, and stylistic preferences.  The
`situation' encodes the information one might find in a traditional
interlingual representation---the definite portion of meaning to be
expressed.  A `possibility' takes as a value a four-part structure of
(1) frequency (never, sometimes, or always), which represents the degree
of possibility; (2) strength (weak, medium, or strong), which represents
how strongly the nuance is conveyed; (3) type (emphasis, suggestion,
implication, or denotation), which represents how the nuance is conveyed;
and (4) an instance of a concept.  The `style' and `attitude'
attributes should be self-explanatory.  As for content, some of the
meanings were discussed in section~\ref{near-syns}, and the rest are
derived from the aforementioned dictionaries.  Comments (labelled with
`\%') are included to indicate which words gave rise to which
possibilities.

\section{Conclusion}

This paper has motivated the need to represent possibilities (or
preferences) in addition to necessary components (or constraints) in
the interlingual representation of a situation.  Possibilities are
required because words can convey a myriad of sometimes indirect
nuances of meaning depending on the context.  Some examples of how one 
could represent possibilities were given.

\section*{Acknowledgements}

For comments and advice, I thank Graeme Hirst.  This work is
financially supported in part by the Natural Sciences and Engineering
Research Council of Canada.

\begin{figure*}
\begin{center}
\begin{avm}
\{situation
     [provide1 \\
      instance-of MakingAvailable \\
      AGENT @{1} [accion-international \\
                  instance-of NonProfitOrganization] \\
      OBJECT [assistance1 \\
              instance-of Helping \\
              ATTRIBUTE [technical1 \\
                         instance-of Technical]] \\
      RECIPIENT @{2} [network \\
                      instance-of Network]] \\

 possibility (frequency sometimes \\
              type suggestion \\
              concept [foresight1 \\
                       instance-of Foreseeing \\
                       AGENT @{1}]) \textit{\% from the word `provides'} \\

 possibility (frequency sometimes \\
              type emphasis \\
              concept [prepare1 \\
                       instance-of Preparing \\
                       AGENT @{1} \\
                       ATTRIBUTE [adequate \\
                                  instance-of Adequacy]]) \textit{\% from `provides'}\\

 possibility (frequency always \\
              type suggestion \\
              concept [subordinate-status \\
                       instance-of Status \\
                       DEGREE [subordinate \\
                               instance-of Subordinate] \\
                       ATTRIBUTE-OF @{1} \\
                       RELATIVE-TO @{2}]) \textit{\% from `assistance'} \}
\end{avm}
\\[2ex]
  ``ACCION International \ldots\ provides technical assistance
  to a network \ldots'' \\
  ``ACCION International \dots\ fournit une assistance technique \`a
  un r\'eseau \ldots'' \\

\caption{Interlingual representation of the `equivalent' sentences shown above.
  Includes four possibilities of what is expressed.}
\label{ex-1}
\end{center}
\end{figure*}

\begin{figure*}
\begin{center}
\begin{avm}
\{situation
     [provide2 \\
      instance-of MakingAvailable \\
      AGENT @{1} [prodem-venture \\
                  instance-of NonProfitJointVenture] \\
      RECIPIENT @{2} [workers \\
                      instance-of Worker] \\
      OBJECT [credit-and-training \\
              instance-of CreditAndTraining \\
              AGENT-OF @{3} [broaden \\
                             instance-of Increasing \\
                             PATIENT @{4} [opportunity \\
                                           instance-of Chance \\
                                           POSSESSED-BY @{2} \\
                                           REGARDING @{5} [employment \\
                                                           instance-of Employment]]]]] \\


 possibility (frequency sometimes \\
              type implication \\
              concept [scope \\
                       instance-of Scope \\
                       MANNER-OF @{3}]) \textit{\% from the word `broaden'}\\

 possibility (type implication \\
              concept [desire \\
                       instance-of Desiring \\
                       AGENT @{2}\\
                       PATIENT @{5}]) \textit{\% from `opportunities'}\\

 possibility (frequency sometimes \\
              strength weak \\
              type suggestion \\
              concept [provoke \\
                       instance-of Provoking \\
                       AGENT @{4} \\
                       PATIENT @{2}]) \textit{\% from `opportunities'} \}
\end{avm}
\\[2ex]
  ``PRODEM \ldots\ provided credit and training to broaden employment
  opportunities \ldots'' \\
  ``PRODEM \dots\ d'offrir \ldots\ des possibilit\'es de cr\'edit et de
  formation pour \'elargir leurs perspectives d'emploi''
\caption{Another interlingual representation with possibilities of
  what is expressed.}
\end{center}
\end{figure*}

\begin{figure*}
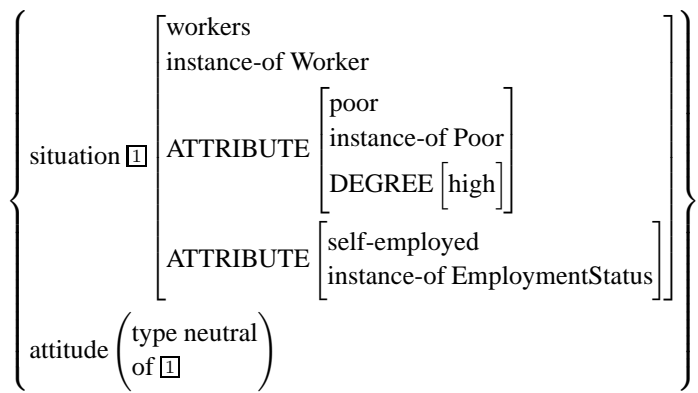

\begin{center}
\begin{avm}
\{ situation 
    @{1} [begin \\
          instance-of Beginning \\
          OBJECT [transition \\
                  instance-of StateChange] \\
          TIME [year-1989 \\
                instance-of Year]] \\
 possibility (type implication \\
              concept [prepare2 \\
                       instance-of Preparing \\
                       AGENT @{1}]) \textit{\% from `amorc\'ee'}  \\

 style (formality (level high)) \}

\end{avm}
\\[2ex]
``The transition \ldots\ began in 1989.'' \\
``La transition, amorc\'ee en 1989 \ldots''
\caption{Interlingual representation with a stylistic preference (for
  high formality).}
\end{center}
\end{figure*}

\begin{figure*}
\begin{center}
\begin{avm}
\{situation
    @{1} [workers \\
          instance-of Worker \\
          ATTRIBUTE [poor \\
                     instance-of Poor \\
                     DEGREE [high]] \\
          ATTRIBUTE [self-employed \\
                     instance-of EmploymentStatus]] \\
 attitude (type neutral \\
           of @{1} ) \}

\end{avm}
\\[2ex]
``the very poor self-employed'' \\
``travailleurs ind\'ependents les plus d'\'emunis''
\caption{Interlingual representation with an expressed attitude.}
\label{ex-last}
\end{center}
\end{figure*}


\bibliographystyle{acl}  

\end{document}